\documentclass[conference]{IEEEtran}
\usepackage{cite}
\usepackage{amsmath,amssymb,amsfonts}
\usepackage{algorithmic}
\usepackage{graphicx}
\usepackage{textcomp}
\usepackage{xcolor}
\usepackage{colortbl}
\usepackage{tikz}
\usepackage[noabbrev,capitalise]{cleveref}
\def\BibTeX{{\rm B\kern-.05em{\sc i\kern-.025em b}\kern-.08em
    T\kern-.1667em\lower.7ex\hbox{E}\kern-.125emX}}
\definecolor{gray1}{gray}{0.88}
\definecolor{gray2}{gray}{0.78}

\newcommand\copyrighttext{\small{978-1-5386-5541-2/18/\$31.00 \textcopyright 2019 IEEE. Personal use of this material is permitted. Permission from IEEE must be obtained for all other uses, in any current or future media, including reprinting/republishing this material for advertising or promotional purposes, creating new collective works, for resale or redistribution to servers or lists, or reuse of any copyrighted component of this work in other works.}}
\newcommand\copyrightnotice{%
\begin{tikzpicture}[remember picture,overlay]
\node[anchor=south,yshift=10pt] at (current page.south) {\fbox{\parbox{\dimexpr\textwidth-\fboxsep-\fboxrule\relax}{\copyrighttext}}};
\end{tikzpicture}%
}

\begin{document}

\title{Identifying Mislabeled Instances in \\Classification Datasets}

\author{\IEEEauthorblockN{Nicolas M. M\"uller}
\IEEEauthorblockA{\textit{Cognitive Security Technologies} \\
\textit{Fraunhofer AISEC}\\
85748 Garching, Germany \\
nicolas.mueller@aisec.fraunhofer.de}
\and
\IEEEauthorblockN{ Karla Markert}
\IEEEauthorblockA{\textit{Cognitive Security Technologies} \\
\textit{Fraunhofer AISEC}\\
85748 Garching, Germany \\
karla.markert@aisec.fraunhofer.de}

}

\maketitle
\copyrightnotice

\begin{abstract}
A key requirement for supervised machine learning is labeled training data, which is created by annotating unlabeled data with the appropriate class.
Because this process can in many cases not be done by machines, labeling needs to be performed by human domain experts.
This process tends to be expensive both in time and money, and is prone to errors. 
Additionally, reviewing an entire labeled dataset manually is often prohibitively costly, so many real world datasets contain mislabeled instances.

To address this issue, we present in this paper a non-parametric end-to-end pipeline to find mislabeled instances in numerical, image and natural language datasets.
We evaluate our system quantitatively by adding a small number of label noise to 29 datasets, and show that we find mislabeled instances with an average precision of more than 0.84 when reviewing our system's top 1\% recommendation.
We then apply our system to publicly available datasets and find mislabeled instances in CIFAR-100, Fashion-MNIST, and others.
Finally, we publish the code and an applicable implementation of our approach.
\end{abstract}


\section{Introduction}
The prediction accuracy of supervised machine learning methods depends on two main ingredients: the quality of the labeled training data and the appropriateness of the algorithm \cite{brodley1999identifying}. 
In this paper, we focus on identifying mislabeled instances in order to improve data quality.
This, in return, may help us to get a higher prediction accuracy when applying a suitable classification algorithm \cite{brodley1999identifying}.

Still today, labeling is either done manually by experts as in Fashion-MNIST \cite{xiao2017fashion} or at least checked by humans as in CIFAR-100 \cite{krizhevsky2009learning}, which costs both time and money.
Errors can occur when labeling is performed by experts as well as when it is performed by non-experts.
\cite{brodley1999identifying} lists subjectivity, data-entry error, and inadequacy of the information as possible causes.
Especially on numerical and image data, the last two are the most important \cite{ekambaram2017finding,alrawi2018labeling}.
For datasets containing as many as 50,000 instances (e.g., CIFAR-100), it is nearly impossible to manually find mislabeled data without additional pre-selection.

In order to address the problem of mislabeled instances, we present a tool set that comprises an end-to-end pipeline to help identify this kind of error. 
The user needs to provide the labeled data to be checked; suitable hyperparameters are inferred automatically.
The tool then returns the instances with the highest probability of carrying a wrong label. 
Hence, it can be used to improve existing datasets as well as to check new datasets before publishing them.
The tool can be applied to any classification problem, whether it is numerical data, images, or natural language.

\begin{figure}[t]
    \centering
    \includegraphics[width=0.11\textwidth]{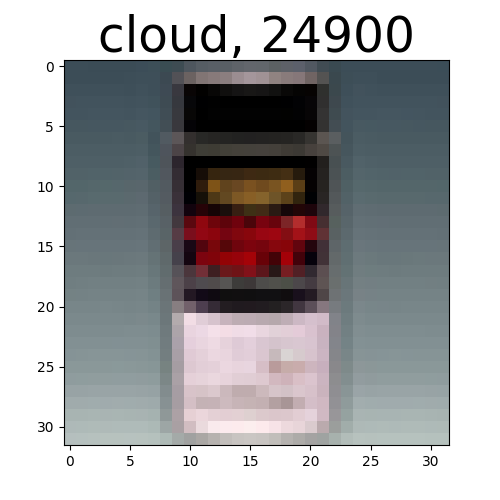}
    \includegraphics[width=0.11\textwidth]{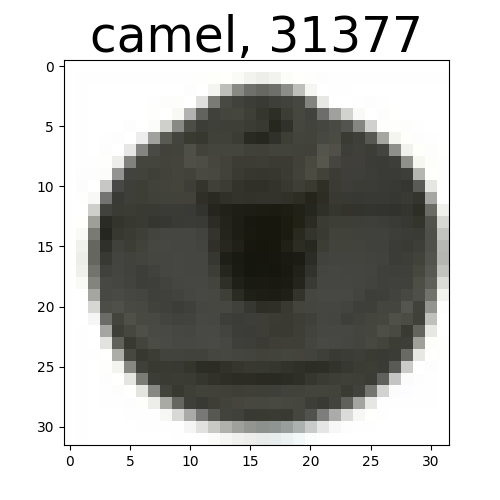}
    \includegraphics[width=0.11\textwidth]{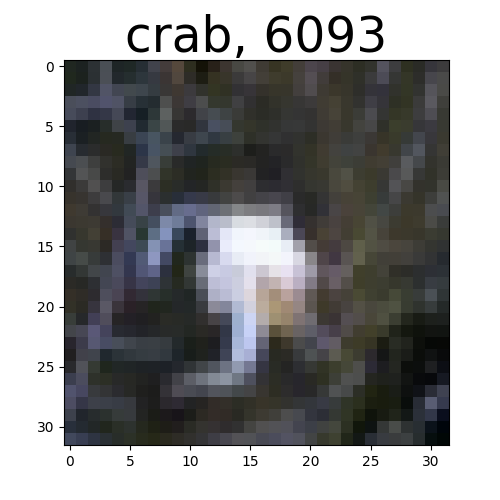}
    \includegraphics[width=0.11\textwidth]{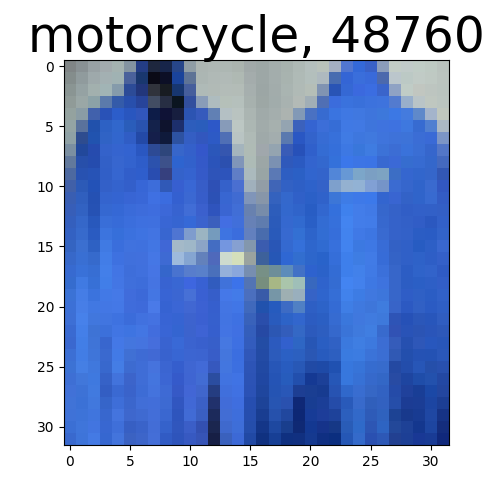}
    \caption{Mislabeled instances in the CIFAR-100 training set, with corresponding label and index of the image in the data set.}
    \label{fig:MislabeledCIFAR}
\end{figure}
\begin{figure}[t]
    \centering
    \includegraphics[width=0.11\textwidth]{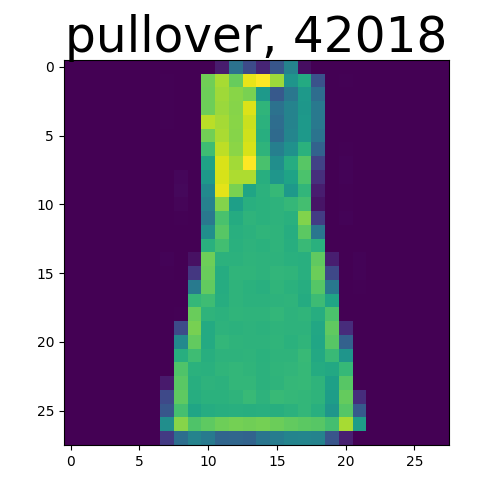}
    \includegraphics[width=0.11\textwidth]{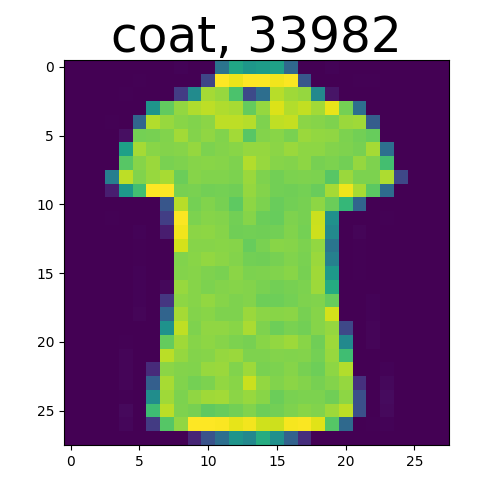}
    \includegraphics[width=0.11\textwidth]{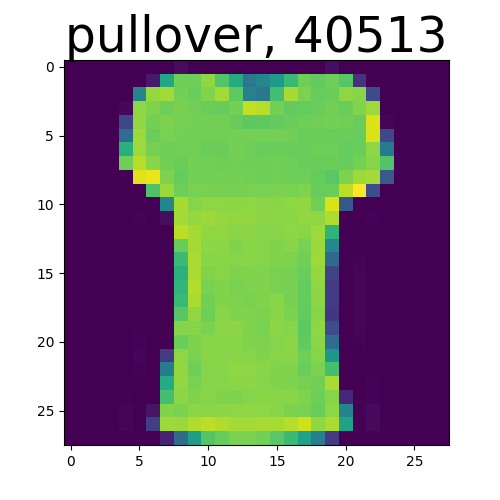}
    \includegraphics[width=0.11\textwidth]{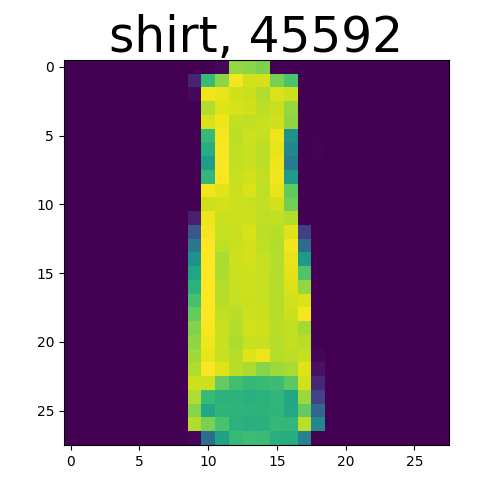}
    \caption{Mislabeled instances in the Fashion-MNIST training set.}
    \label{fig:MislabeledFashionMNIST}
\end{figure}

For the empirical evaluation of our tool set, we use a combination to 29 real-world and synthetic datasets, among these the famous MNIST, CIFAR, Twenty Newsgroup and IMDB datasets. 
We show that our approach can successfully identify mislabeled instances with label noise both completely at random (independent of the class) and at random (where some classes might be confused more easily) \cite{frenay2014comprehensive,frenay2014classification}.
Our experiments show that the tool set is applicable to a large variety of classification datasets.

In summary, our contribution is as follows.
\begin{itemize}
    \item We provide a tool to automatically pre-select instances likely to be mislabeled, that is suitable for use by non-machine learning practitioners due to its complete independence of user-supplied hyperparameters.
    \item We provide a thorough evaluation of our approach with 29 different real-world and synthetic datasets.
    \item We identify mislabeled instances in CIFAR-100, MNIST and Fashion-MNIST, that, to the best of our knowledge, have not been published before.
    \item We supply the full source code, including the scripts to create the qualitative and quantitative evaluation\footnote{https://github.com/mueller91/labelfix}.
\end{itemize}

This paper is structured as follows. In \cref{sec:RelatedWork} we present related work on label noise detection, followed by a description of our method in \cref{sec:Methodology}. 
The empirical evaluation on both artificial and real-world datasets is given in \cref{sec:EmpiricalEvaluation}. 
Finally, \cref{sec:Conclusion} and  \cref{sec:FutureWork} conclude the paper with an outlook on possible topics for further research.

\section{Related Work}\label{sec:RelatedWork}
There are three ways to deal with noisy datasets. First, it is possible to design robust algorithms that can learn even from noisy data. 
Second, mislabeled instances can be found and automatically removed from the dataset before training. 
Third, mislabeled instances can be identified and then re-evaluated by a domain expert. 
In this section, we briefly present related work in each of these categories.

\subsection{Learning with label noise}\label{subsec:RelatedLabelNoise}
A comprehensive overview on class label noise is provided by \cite{frenay2014comprehensive,frenay2014classification}, which summarize the literature on label noise and distinguish three sources of noise: completely at random, at random, and not at random. 
According to their definition, the first kind of labeling error occurs independently of the true class and the feature values (hence, anything can possibly be confused with anything else). 
The second kind of error only depends on the class (e.g., confusing lions and leopards but not lions and trees), whereas the third class, however, depends on the class as well as the feature values (e.g., confusing only big yellowish cats with lions but not black ones).

As pointed out by \cite{frenay2014classification}, the impact of label or class noise is generally higher than feature noise because there is only one label but many features, some of which may be redundant.

In, e.g., \cite{li2017learning,bootkrajang2016generalised} label noise robust algorithms are introduced. 
Another approach is presented in \cite{hendrycks2018using}, which proposes a correction method that uses trusted data, i.e., data labeled by experts, to improve robustness to label noise. 
The technique is applied to artificial modifications of the MNIST, the CIFAR-10, the CIFAR-100, the IMDB Large Movie Reviews, the Twitter Part of Speech, and the Stanford Sentiment Treebank dataset.
As even experts make labeling mistakes (see \cref{sec:EmpiricalEvaluation}), we focus on a general approach that does not require additional knowledge and handles both completely at random and random class noise.

\subsection{Label cleansing}\label{subsec:RelatedDataCleansing}
In \cite{brodley1999identifying}, the authors propose to first train filters on parts of the training data in order to identify mislabeled examples in the remaining training data.
To this end, they use cross-fold validation among decision trees, nearest neighbor classifiers and linear machines. 
Mislabeled instances are subsequently removed from the training set.
Then, a learning algorithm is trained on the reduced training set. 
They show empirically that feeding a supervised algorithm with filtered instances results in a higher predictive accuracy than without previous data cleansing.

In \cite{sabzevari2018two}, random forests are trained on bootstrapped samples in order to create ensembles. 
A threshold for the disagreement ratio of the classifiers is introduced to detect mislabeled data on artificially disturbed datasets such as breast\_cancer and liver.

\subsection{Label noise identification}\label{subsec:RelatedLabelNoiseIdent}
Recent publication \cite{ekambaram2017finding} makes use of a two-level approach that requires human interaction: First, a support vector machine (SVM) is trained on the unfiltered data. 
Then, a second binary classifier, which is used for label prediction, is trained on the original data without the support vectors. 
If the original label and the predicted label do not match, it is a possible label noise that will then be presented to a human checker.
This method is applied to pairs of classes in ImageNet that can be easily confused.
After reviewing over 15\% of the images in ImageNet, the authors could identify 92 mislabeled pictures.

Another promising approach is presented in \cite{alrawi2018labeling}. The authors train a convolutional neural network (CNN) ensemble on parts of the dataset in order to calculate predictions for the remaining dataset based on majority voting.  
This method is applied to different image datasets (CIFAR10, CIFAR100, EMNIST, and SVHN).
For a comparison to our results, see \cref{subsec:improve}.

Although most publications have focused on numerical and image data, where errors are easier to spot by human, similar procedures can be applied to, e.g., text documents \cite{esuli2013improving} and e-mail spam filtering \cite{kolcz2009genre}.
In this context however, the problem of ambiguity becomes more prominent, meaning that even for a domain expert, it is not always clear what the correct label should be.

\section{Methodology}\label{sec:Methodology}
The following section details how we find mislabeled instances in a given dataset $\mathcal{D} = (x, y)$.
We assume the matrix $x$ to have shape $(N, D)$, i.e., to hold $N$ data instances each with dimension $D$.
The matrix $y$ has shape $(N, C)$, holding the $N$ data labels in one-hot encoded format, which for $C$ classes yields vectors of length $C$. 
Each instance is assigned exactly one class label.

We define $\mathcal{I} \subseteq \mathcal{D}$ as the set of mislabeled instances.
Hence, $\mathcal{I} = \emptyset$ if there are no mislabeled instances in $\mathcal{D}$.
The set $\mathcal{I}$ is, of course, unknown to the user but for any given $n \in \{0, \, \ldots, \, N-1\}$ a domain expert can manually check whether $\left(x_n, \, y_n \right) \in \mathcal{I}$.
We refer to some domain expert reviewing every single $d \in \mathcal{D}$ as the \emph{naive approach}.
As detailed in the introduction, this extensive re-checking is often prohibitively expensive. 
We assume that $\vert \mathcal{I} \vert \ll N$, because otherwise the naive approach is sufficient.

In order to reduce the number of required reviews, our goal is to construct a mapping $f_\alpha: \mathcal{D} \to \mathcal{I}_\alpha$ for some $0 < \alpha < 1$ (determining the number of instances to be reviewed) such that $| \mathcal{I}_\alpha \cap \mathcal{I} |$ is maximized while $|\mathcal{I}_\alpha|$ is minimized.
Here, $\mathcal{I}_\alpha$, with $|\mathcal{I}_\alpha| = \alpha N$, is the set of instances which are to be reviewed by a domain expert. 
Thus, we aim to construct a system $f_\alpha$ that suggests a set of potentially mislabeled instances $\mathcal{I}_\alpha$.
Our goal is to have the set $\mathcal{I}_\alpha$ contain as many truly mislabeled instances as possible.

In order to make the system user-friendly, the only parameters to be supplied by the user are $\alpha$ (determining the size of the output set) and the labeled dataset $\mathcal{D}$.

\subsection{System pipeline}\label{subsec:system_pipeline}
In order to detect mislabeled instances in a given dataset $\mathcal{D}$, we propose the following pipeline.
\begin{enumerate}
    \itemsep0em
    \item A dataset $\mathcal{D} = (x, y)$ and a parameter $\alpha$ are supplied by the user. Here, $\alpha$ is the percentage of images the user is willing to manually re-evaluate.
    \item Based on the shape of $x$, the corresponding network layout and best hyperparameters are found automatically. See \cref{sec:model_selection} for details.
    \item Preprocess the data and train a model $g$ on the dataset $(x, y)$.
    \item Use the trained model $g$ to re-classify $x$, and obtain $g(x) = \Tilde{y}$.
    \item For all $n \in \{0, \, \ldots, \, N-1\}$, the inner product $\langle y_n, \Tilde{y_n} \rangle$ is calculated, which yields the probability that instance $x_n$ gets assigned the original label $y_n$.
    \item Finally, the set $\mathcal{I}_{\alpha} \subset \{0, ..., N-1\}$ is returned, such that $ \sum_{\mathcal{I}_\alpha} \langle y_i, \Tilde{y_i} \rangle $ is minimized and $|\mathcal{I}_\alpha| = \alpha N$.
    This is achieved by first sorting $\langle y_n, \Tilde{y_n} \rangle_{n \in \left\{0, ..., N-1\right\}} $ by argument in ascending order and then returning the first $\alpha N$ instances\footnote{Alternatively, the user may choose $\alpha = 1$ and review the returned, ordered instances until their time or money resources are depleted.
    Since the probability of being mislabeled decreases with every instance, this will optimally utilize available resources.}.
\end{enumerate}
In summary, we train a classifier $g$ on a given dataset $(x, y)$, and use the same classifier $g$ to obtain class probabilities for $x$.
We then look for instances for which the class probability of the original label is minimal, e.g., instances for which the classifier considers the original label extremely unlikely given the feature distribution learned during training.
These instances are returned to the user for re-evaluation.

\subsection{Data preprossessing}\label{subsec:DataPrep}
As with all machine learning pipelines, data preprossessing is an essential step in our system. We preprocess numerical, image and natural language data as follows.
\begin{itemize}
    \itemsep0em 
    \item Numerical data is preprocessed by feature-wise scaling to $[0, 1]$.
    This process is known as \emph{MinMax scaling}.
    \item Image data is preprocessed by feature-wise setting the data mean to 0 and feature-wise dividing by the standard deviation (std) of the dataset. This process is known as \emph{standardization}.
    \item Natural language data is preprocessed using word embeddings \cite{w2v}
    as follows: First, we apply very basic textual preprocessing such as splitting on whitespaces, then we map individual words to 300-dimensional word embeddings using a pre-trained embedding \cite{eyalerwo16:online},
    and finally we retrieve the corresponding representation for the whole sequence of words by simply summing up the individual embeddings.
\end{itemize}

\subsection{Classification algorithm}\label{sec:model_selection}

\begin{table}
    \centering
    \caption{Parameter grid for cross-validation based hyperparameter optimisation. For Image data, we use a convolutional network with fixed hyper parameters.}
    \begin{tabular}{|c|c|c|c|}
        \hline
                          & \textbf{Numerical}     &   \cellcolor{gray2} \textbf{Textual}       \\ \hline
        Depth             & $\left\{ 1, \, 2, \, 3, \, 5 \right\}$ &     \cellcolor{gray2}$\left\{ 1, \, 2, \, 3, \, 5 \right\}$ \\ 
        Units per Layer   & $\left\{ 50, 120 \right\}$    &     \cellcolor{gray2}$\left\{ 50, 120 \right\}$    \\ 
        Dropout per Layer & $\left\{ 0, \, 0.1, \, 0.2 \right\}$    &   \cellcolor{gray2}$\left\{ 0, \, 0.1, \, 0.2 \right\}$   \\ 
        \hline
        \end{tabular}
    \label{tab:Models}
\end{table}

In order to find mislabeled instances in a dataset as described in \cref{subsec:system_pipeline}, a robust classifier $g$ is required.
That is to say, $g$ must be able to generalize well and not overfit the dataset $\mathcal{D}$.
This is necessary because since $\mathcal{D} = (x, y)$ is assumed to be noisy (due to mislabeled instances), overfitting on $\mathcal{D}$ results in the classifier simply `remembering' all instances $(x_n, y_n)$ for $n \in \{0, \, \ldots, \, N-1\}$.
In this setting, deviations in individual $(x_n, y_n)$ from the overall distribution would not be found, thus severely diminishing the system's performance.

Hence, $g$ needs to 1) be flexible to correctly learn from a variety of text, image and numerical datasets and 2) generalize well on noisy datasets.
Neural networks, especially with dropout layers are a natural choice in this setting.
How we apply them to the individual categories of datasets is examined in the following subsection.

\subsection{Automatic hyperparameter selection}\label{subsec:hyper_param_sel}
We aim to make finding mislabeled instances as easy as possible. 
Hence, our system automatically selects a suitable classifier and corresponding hyperparameters according to the dataset. 
This enables scientists from non-machine learning domains to easily examine their own data.
We find suitable network architectures and hyperparameters for each dataset type as follows.

\begin{itemize}
    \item For numerical and textual data, we use a dense feed forward neural network and tune the following hyperparameters: number of hidden layers, number of units per hidden layers, and dropout.
    We use 3-fold cross-validation to search the predefined hyperparameter space (see \cref{tab:Models}) for the best combination of parameters. For preprocessing, we apply data normalization. Additionally, textual data is mapped to vector representation using word embeddings. 
    \item For image data, we use a convolutional network. It consists of three blocks: Block one consists of two convolutional layers with 48 2x2 kernels, followed by 3x3 maxpooling and 25\% dropout. Block two is the same as block one, but with 96 kernels. Block three flattens the convolutional output of block two and applies three dense layers with 50\% Dropout and ReLu nonlinearities. 
    We chose this architecture because convolutions work for all kinds of input images, as convolutional layers can cope with varying image input sizes.
    Note that we omit the hyperparameter grid search in the interest of computation time.
    Overfitting is prevented using very high dropout (between 25 and 50\% per layer) and early stopping on the validation accuracy.
    We also experimented with transfer learning, using a ResNet50 as feature extractor, followed by a dense network. 
    We find that this approach is inferior to our convolutional network in terms of precision and recall.
\end{itemize}
We chose these architectures since they are standard baselines which have been shown to work with the corresponding data type \cite{lecun, leskovec2014mining, yosinski2014transferable}.
Since we want the networks to be able to generalize well and learn even from noisy data, we need to prevent the networks from overfitting, i.e., simply "remembering" the training data.
Thus, we select the hyperparameters via cross-validated grid search.
Additionally, we add several levels of drop-out to the parameter grid and use an aggressive learning rate of 1e-2, which also cuts computation time. Also, we use early stopping with \emph{patience}$=15$ and \emph{threshold}$=0.005$.

Finally, in order to deal with imbalanced class labels, we adjust the gradients during training  using  class  weights and  employ balanced  F-score  as  an  evaluation  metric.

\section{Empirical Evaluation}\label{sec:EmpiricalEvaluation}
We evaluate our system both qualitatively by finding mislabeled instances in well-researched datasets such as CIFAR-100 and Fashion-MNIST, as well as quantitatively by artificially constructing mislabeled instances ourselves and identifying them with our tool.
In the following section, we describe this evaluation in detail and present results.
Note that we make all experiments accessible by publishing our source code.

\subsection{Quantitative evaluation}\label{subsec:quant_eval}
In order to evaluate the effectiveness of our system in a quantitative manner, we propose the following.
\begin{enumerate}
    \itemsep0em
    \item We select 29 datasets from the domain of text, image and numerical data, for each of which we flip\footnote{That is to say, deliberately assign an incorrect label.} $\mu$ percent of the labels. We refer to this new dataset as $\mathcal{D}'$.
    \item We apply our tool set to this noisy dataset $\mathcal{D}'$ as described in \cref{subsec:system_pipeline}. 
    \item We evaluate the results of our approach with respect to different metrics by comparing the system's output $\mathcal{I}_\alpha$ to the actual set of flipped labels.
\end{enumerate}
We use the following datasets (see also \cref{tab:table2}).
\begin{itemize}
    \item From \verb|sklearn| \cite{pedregosa2011scikit}: breast cancer, digits, forest covertype, iris, twenty newsgroup, and wine;
    \item from \verb|kaggle.com|: pulsar-stars, sloan digital sky survey
    \item from the \verb|UCI Machine Learning Repository| \cite{dua2017UCI}: adult, credit card default \cite{yeh2009comparisons}, SMS spam \cite{almeida2011contributions}, and liver;
    \item from \verb|keras| \cite{chollet2015keras}: CIFAR-10 \cite{krizhevsky2009learning}, CIFAR-100 \cite{krizhevsky2009learning}, Fashion-MNIST \cite{xiao2017fashion}, IMDb \cite{maas2011learning}, MNIST \cite{deng2012mnist}, and SVHN \cite{netzer2011reading}.
\end{itemize}
Additionally, we have created six synthetic datasets with varying feature and class sizes (see \cref{tab:table2}) using  \verb|sklearn|'s 
\verb|make_classification| plus one $blob$ dataset using \verb|sklearn|'s \verb|make_blobs|.

\begin{table}[]
    \centering
    \caption{Overview of the datasets.}
    \begin{tabular}{|l|c|c|r|}
\hline
\textbf{Dataset} & \textbf{Size} & \textbf{Type} & \textbf{Classes}\\
\hline
adult &         (32561, 14) &  numerical &        2 \\
breast\_cancer &           (569, 30) &  numerical &        2 \\
\rowcolor{gray1} cifar10 &  (50000, 32, 32, 3) &      image &       10 \\
\rowcolor{gray1}cifar100 &  (50000, 32, 32, 3) &      image &      100 \\
\rowcolor{gray1}cifar100, at random &  (50000, 32, 32, 3) &      image &      100 \\
\rowcolor{gray1}cifar100, subset aqua &   (2500, 32, 32, 3) &      image &        5 \\
\rowcolor{gray1}cifar100, subset flowers &   (2500, 32, 32, 3) &      image &        5 \\
\rowcolor{gray1}cifar100, subset household &   (2500, 32, 32, 3) &      image &        5 \\
credit card default &         (30000, 23) &  numerical &        2 \\
digits &          (1797, 64) &  numerical &       10 \\
\rowcolor{gray1}fashion-mnist &  (60000, 28, 28, 3) &      image &       10 \\
forest covertype (10\%) &         (58101, 54) &  numerical &        7 \\
\rowcolor{gray2}imdb &        (25000, 100) &    textual &        2 \\
iris &            (150, 4) &  numerical &        3 \\
\rowcolor{gray1}mnist &  (60000, 28, 28, 3) &      image &       10 \\
pulsar\_stars &          (17898, 8) &  numerical &        2 \\
sloan-digital-sky-survey &         (10000, 17) &  numerical &        3 \\
\rowcolor{gray2}sms spam &         (5572, 300) &    textual &        2 \\
\rowcolor{gray1}svhn &  (73257, 32, 32, 3) &      image &       10 \\
synthetic 1 &          (10000, 9) &  numerical &        3 \\
synthetic 2 &          (10000, 9) &  numerical &        5 \\
synthetic 3 &         (10000, 45) &  numerical &        7 \\
synthetic 4 &         (10000, 45) &  numerical &       15 \\
synthetic 5 &         (10000, 85) &  numerical &       15 \\
synthetic 5 &         (10000, 85) &  numerical &        7 \\
synthetic blobs &          (4000, 12) &  numerical &       12 \\
\rowcolor{gray2}twenty newsgroup &        (18846, 300) &    textual &       20 \\
\rowcolor{gray2}twitter airline &        (14640, 300) &  textual &        3 \\
wine &           (178, 13) &  numerical &        3 \\
\hline
\end{tabular}

    \label{tab:table2}
\end{table}

Estimating the ratio of mislabeled instances $\mu$ for real-world datasets is somewhat arbitrary and may differ significantly from case to case. However, \cite{frenay2014comprehensive} suggests that real-world datasets have around 5\% mislabeled data.
We present our results for $\mu = 0.03$ in \cref{tab:table3}, as we believe a 5\% error rate to be too high for these well-researched datasets. Results for other choices of $\mu$ are comparable and may easily be created by re-running the provided script with different parameters.

In order to simulate noise at random and noise completely at random, we have introduced class-independent label noise on all datasets as well as class-dependent label noise in CIFAR-100 (see \cref{tab:table3}, "cifar100, at random``). Here, only labels within the same subgroup are interchanged.

We report two different goodness criteria: the $\alpha$-precision and the $\alpha$-recall, defined by
\begin{align}
    \alpha\text{-precision} &= \frac{\vert \mathcal{I}_\alpha \cap \mathcal{I} \vert}{\vert \mathcal{I}_\alpha 
    \vert},\label{eq:acc} \\
    \alpha\text{-recall} &= \frac{\vert \mathcal{I}_\alpha \cap \mathcal{I} \vert}{ \vert \mathcal{I} \vert}.\label{eq:good}
\end{align}

\begin{table*}[ht]
    \centering
    \caption{Precision and recall values for artificially added 3\% noise, averaged over five runs.}
    \begin{tabular}{|l|r|rrr|rrr|}
\hline
\textbf{Dataset} & \textbf{Runtime} & \multicolumn{3}{|c|}{\textbf{$\alpha$-precision}} & \multicolumn{3}{|c|}{\textbf{$\alpha$-recall}}\\
& & \multicolumn{1}{|c}{$\alpha = 0.01$} & \multicolumn{1}{c}{$\alpha = 0.02$} & \multicolumn{1}{c|}{$\alpha = 0.03$} & \multicolumn{1}{|c}{$\alpha = 0.01$} & \multicolumn{1}{c}{$\alpha = 0.02$} & \multicolumn{1}{c|}{$\alpha = 0.03$} \\
\hline

adult &    2.1 min &       0.80 &       0.63 &       0.51 &      0.27 &      0.42 &      0.51 \\
breast\_cancer &   34.0 sec &       0.76 &       0.80 &       0.74 &      0.22 &      0.52 &      0.74 \\
\rowcolor{gray1}cifar10 &   9.47 min &       0.98 &       0.88 &       0.72 &      0.33 &      0.59 &      0.72 \\
\rowcolor{gray1}cifar100 &  13.07 min &       0.94 &       0.82 &       0.67 &      0.31 &      0.54 &      0.67 \\
\rowcolor{gray1}cifar100, at random &  11.48 min &       0.43 &       0.35 &       0.31 &      0.14 &      0.23 &      0.31 \\
\rowcolor{gray1}cifar100, subset aqua &   20.2 sec &       0.61 &       0.38 &       0.32 &      0.20 &      0.25 &      0.32 \\
\rowcolor{gray1}cifar100, subset flowers &   32.8 sec &       0.63 &       0.43 &       0.34 &      0.21 &      0.29 &      0.34 \\
\rowcolor{gray1}cifar100, subset household &   48.6 sec &       0.62 &       0.46 &       0.37 &      0.21 &      0.30 &      0.37 \\
credit card default &    1.9 min &       0.18 &       0.17 &       0.18 &      0.06 &      0.12 &      0.18 \\
digits &   51.8 sec &       0.98 &       0.95 &       0.86 &      0.31 &      0.63 &      0.86 \\
\rowcolor{gray1}fashion-mnist &  10.71 min &       0.99 &       0.98 &       0.90 &      0.33 &      0.66 &      0.90 \\
forest covertype (10\%) &    4.6 min &       1.00 &       0.95 &       0.74 &      0.33 &      0.63 &      0.74 \\
\rowcolor{gray2}imdb &   3.71 min &       0.70 &       0.61 &       0.51 &      0.23 &      0.41 &      0.51 \\
iris &   26.9 sec &       1.00 &       0.53 &       0.55 &      0.25 &      0.40 &      0.55 \\
\rowcolor{gray1}mnist &   3.74 min &       1.00 &       1.00 &       0.97 &      0.33 &      0.67 &      0.97 \\
pulsar\_stars &   51.8 sec &       0.91 &       0.86 &       0.78 &      0.30 &      0.57 &      0.78 \\
sloan-digital-sky-survey &    1.5 min &       0.80 &       0.71 &       0.63 &      0.27 &      0.47 &      0.63 \\
\rowcolor{gray2}sms spam &   1.44 min &       0.85 &       0.86 &       0.79 &      0.28 &      0.57 &      0.79 \\
\rowcolor{gray1}svhn &   13.6 min &       0.92 &       0.90 &       0.83 &      0.31 &      0.60 &      0.83 \\
synthetic 1 &   2.05 min &       1.00 &       0.98 &       0.89 &      0.33 &      0.66 &      0.89 \\
synthetic 2 &   2.74 min &       1.00 &       0.99 &       0.89 &      0.33 &      0.66 &      0.89 \\
synthetic 3 &   3.79 min &       1.00 &       0.99 &       0.91 &      0.33 &      0.66 &      0.91 \\
synthetic 4 &    4.9 min &       0.98 &       0.90 &       0.74 &      0.33 &      0.60 &      0.74 \\
synthetic 5 &   3.53 min &       0.95 &       0.84 &       0.70 &      0.32 &      0.56 &      0.70 \\
synthetic 6 &   3.58 min &       1.00 &       0.98 &       0.86 &      0.33 &      0.65 &      0.86 \\
synthetic blobs &   37.8 sec &       1.00 &       1.00 &       0.98 &      0.33 &      0.67 &      0.98 \\
\rowcolor{gray2}twenty newsgroup &    3.2 min &       0.79 &       0.73 &       0.63 &      0.26 &      0.49 &      0.63 \\
\rowcolor{gray2}twitter airline &   2.39 min &       0.66 &       0.52 &       0.43 &      0.22 &      0.34 &      0.43 \\
wine &   28.1 sec &       1.00 &       1.00 &       0.88 &      0.20 &      0.60 &      0.88 \\
\hline
\textbf{Averages} &         &       0.84 &       0.77 &       0.68 &      0.27 &      0.51 &      0.68 \\
\hline
\end{tabular}

    \label{tab:table3}
\end{table*}

The $\alpha$-precision (see \cref{eq:acc}) returns the ratio of the number of flipped labels among the returned $\alpha N$ instances to the output size $\alpha N$ (i.e., how many of the system's suggestions are indeed wrongly labeled?).
For example, an $\alpha$-precision value of $0.8$ for $\alpha = 0.01$ and $\mu = 0.03$ for some dataset $\mathcal{D}$ specifies the following. Among the set $\mathcal{I}_\alpha$, which has size $0.01N$, we found $80\%$ flipped labels. Hence, this method is more than 26 times better than random guessing, but still returns $20\%$ correctly labeled instances to be re-checked.

The $\alpha$-recall (see \cref{eq:good}) specifies how many of the flipped labels are found by looking only at the first $\alpha$ percent of the data (i.e., how many of the wrongly labeled instances are present in the system's output suggestions?).
For example, assume the following:
Let $\alpha=0.03$ and $\mu = 0.03$ for some dataset $\mathcal{D}$. A recall value of $0.8$ specifies that we find $80\%$ of the $3\%$ flipped labels if we review only $0.03 N$ instances as suggested by our system.

As the returned instances are already sorted by their probability to be mislabeled, we expect the $\alpha$-precision to become lower for higher values of $\alpha$ as less new mislabeled instances are added to $\mathcal{I}_{\alpha}$. On the other hand, the $\alpha$-recall increases for higher values of $\alpha$, as $\alpha = 1$ implies that $\mathcal{I}_\alpha \subseteq \mathcal{I}$, hence $\alpha\text{-recall}=1$. Furthermore, for $\alpha = \mu$ we have $\alpha-\text{precision} = \alpha-\text{recall}$ (see \cref{tab:table3} for $\alpha = \mu = 0.03$).

We run our tool on 29 datasets with $\alpha=0.03$, and present the results in \cref{tab:table3}.
They indicate the following.
\begin{enumerate}
    \item Our tool works equally well on the different data types tested (i.e., images, natural language, and numerical data).
    \item This method detects severe, obvious labeling errors among very different classes (e.g., bridge and apple) as well as less obvious errors among rather similar classes (e.g., roses and tulips from the CIFAR-100 subset flowers, see "cifar-100, at random``),
    \item The algorithm has reasonable computing time, depending on the dataset size and the number of classes. We use an Intel E5-2640 based machine and assigned 20 cores to perform our computations.
    The convolutional network is trained on an Nvidia Titan X GPU.
\end{enumerate}
In comparison to \cite{alrawi2018labeling}, we are training only one neural network, which saves us time while achieving very good results (see \cref{tab:table3}). 
Furthermore, our approach is more general as, e.g., natural language can be processed as well. 
As opposed to \cite{ekambaram2017finding}, we do not require additional information such as classes that are easily confused.

\subsection{Qualitative evaluation}
For the quantitative evaluation in \cref{subsec:quant_eval}, we have only considered our intentionally introduced label noise of $3\%$ (assuming the datasets to be noise-free). During our qualitative evaluation, we apply our tool set to selected datasets without adding extra noise. The datasets we consider in this section contain either images or natural language, since we can easily spot mislabeled instances in these domains.

With $\alpha = 0.003$, we manually review 150 images in CIFAR-100 and 180 images in Fashion-MNIST. In doubt, we stick to the assigned label. Following this procedure, we are able to detect some errors that, to the best of our knowledge, have not been reported yet. Our findings include the following mislabeled instances.
\begin{itemize}
    \item CIFAR-100: 24900, 31377, 48760; for illustrating pictures, see \cref{fig:MislabeledCIFAR}; for more examples, refer to \cref{tab:CIFARmislabeled};
    \item Fashion-MNIST: 40513 (pullover $\rightarrow$ top/shirt), 33982 (coat $\rightarrow$ shirt), 42018 (pullover $\rightarrow$ dress); for illustrating pictures, see \cref{fig:MislabeledFashionMNIST}; for more examples, refer to \cref{tab:FashionMNISTmislabeled}.
\end{itemize}
In order to show that our tool can also process natural language, we have applied it to the twenty newsgroup dataset, in which each e-mail is assigned one label. 
As all messages are labeled according to the newsgroup they were published in \cite{cmu}, these instances can technically not be considered mislabeled.
Still, we show that our method detects instances for which the class label and the e-mail content do not fit, see \cref{tab:twenty_newsgroups1} and \cref{tab:twenty_newsgroups2}.

The above listing is not supposed to be complete but rather to illustrate that 1) the application of the tool set is indeed easily possible even for larger datasets and 2) even well-researched datasets may contain many mislabeled instances. 
Hence, a semi-automated tool to simplify the process of re-checking is of great importance.

\subsection{Improvement to related work}\label{subsec:improve}
With our tool set, we overcome a fundamental restriction imposed by \cite{ekambaram2017finding}, who require a set of classes to be easily confused.
Additionally, as they make use of an SVM, they can only compare two classes that are commonly confused.
By definition of our algorithm, any two classes can be regarded as possibly confused.
Finally, besides detecting noise at random, our tool set can also identify noise completely at random. 

In comparison to \cite{alrawi2018labeling}, we are more rigorous in considering something as "mislabeled`` (see \cref{sec:appendix}). 
Furthermore, we show that by training only a single architecture instead of an ensemble, comparable and possibly better results can be achieved with less overhead.
Additionally, we require no test-train split and can therefore process smaller datasets more accurately.
The biggest improvement though is that our tool is not limited to image data but can be applied to purely numerical data and natural language data as well.

\section{Conclusion}\label{sec:Conclusion}
In this paper, we presented an automated system to help finding potentially mislabeled instances in classification datasets.
This has the potential to improve classifiers trained on these datasets, especially when the datasets have, unlike CIFAR, not been exposed to a large audience and can only be reviewed by a few domain experts.
Since we cannot assume these domain experts to be proficient in machine learning techniques, we designed our system such that no hyperparameters have to be supplied -- the system works out of the box for numerical, image and text datasets.
The only input data required is the dataset to be analyzed and a parameter that identifies the size of the output set.
Hyperparameters are inferred automatically in order to increase the tool's usability.

This simple label cleansing method can detect noise both among different as well as among similar classes.
We evaluated our system on over 29 datasets on which we added a small fraction of label noise, and found mislabeled instances with an averaged precision of 0.84 when reviewing our system's top 1\% recommendation.
Applying our system to real-world datasets, we find mislabeled instances in CIFAR-100, Fashion-MNIST and others (see Appendix).

\section{Future work}\label{sec:FutureWork}
While our results are promising, we identify the following future work.
First, a generalization of our framework to multi-labeled datasets is desirable. 
To this end, the implemented sigmoid function must be generalized to independent class probabilities. 
Then, instead of using the dot product, one needs to consider the probabilities of the original labels independently.
The general pipeline may remain the same.

Secondly, we would like to measure the impact of our system with respect to the performance of state-of-the-art classifiers. According to our qualitative evaluation, even well-researched datasets contain label noise. Hence, the best performing algorithms that nowadays have error rates below estimated label noise ratios (see, e.g., \cite{alvear2018improving} for MNIST and \cite{cubuk2018autoaugment} for CIFAR-10) should be re-evaluated on the cleaned training and especially testing data. 

Thirdly, other architectures for finding mislabeled data instances should be explored. Another promising method to detect mislabeled instances seems to be the combination with outlier analysis, as, e.g., kernel density estimation. Special attention should be paid to high dimensional data \cite{zimek2012survey}.
Design and evaluation of such a system in comparison to the system presented in this paper might be interesting.

Finally, it would be desirable to have a theoretical comparison to two-level approaches such as \cite{ekambaram2017finding} and voting schemes \cite{alrawi2018labeling}.

\bibliographystyle{unsrt}
\bibliography{manuscriptBib}

\clearpage
\section{Appendix}\label{sec:appendix}
Here, we present some of our findings when applying our system to CIFAR100, MNIST, Fashion-MNIST and the Twenty Newsgroup dataset.
\begin{table}[htbp]
    \centering
    \caption{An example of an instance from the twenty newsgroup dataset where the content does not meet the label.}
    \begin{tabular}{|p{0.45\textwidth}|}
\hline
Index Nr. 13622 \\ \hline
Category: sci.med \\ \hline
From: turpin@cs.utexas.edu (Russell Turpin)\\ 
Subject: Meaning of atheism, agnosticism  (was: Krillean Photography)\\
\\ 
Sci.med removed from followups.  (And I do not read any of the other newsgroups.)\\ 
\\ 
\textgreater As a self-proclaimed atheist my position is that I \_believe\_ that \\ 
\textgreater there is no god.  I don't claim to have any proof.  I interpret\\
\textgreater the agnostic position as having no beliefs about god's existence.\\ 
\\ 
As a self-proclaimed atheist, I believe that *some* conceptions of god are inconsistent or in conflict with fact, and I lack belief in other conceptions of god merely because there is no reason for me to believe in these.  I usually use the word agnostic to mean someone who believes that the existence of a god is unknown inherently unknowable. Note that this is a positive belief that is quite different from not believing in a god; I do not believe in a god, but I also do not believe the agnostic claim. \\ 
\hline
\end{tabular}
    \label{tab:twenty_newsgroups1}
\end{table}
\begin{table}[htbp]
    \centering
    \caption{A second example of an instance from the twenty newsgroup dataset where the content does not meet the label.}
    \begin{tabular}{|p{0.45\textwidth}|}
\hline
Index Nr. 13203 \\ \hline
Category: talk.politics.mideast \\ \hline
From: kevin@cursor.demon.co.uk (Kevin Walsh) \\
Subject: Re: To All My Friends on T.P.M., I send Greetings \\
Reply-To: Kevin Walsh $<$kevin@cursor.demon.co.uk$>$ \\
Organization: Cursor Software Limited, London, England \\
Lines: 17 \\
\\
In article OAF.93May11231227@klosters.ai.mit.edu oaf@zurich.ai.mit.edu writes: \\
\textgreater In message: C6MnAD.MxD@ucdavis.edu Some nameless geek \\ szljubi@chip.ucdavis.edu writes: \\
\textgreater \textgreater To Oded Feingold: \\
\textgreater \textgreater  \\
\textgreater \textgreater Call off the dogs, babe. It's me, in the flesh. And no, I'm not \\
\textgreater \textgreater Wayne either, so you might just want to tuck your quivering erection \\
\textgreater \textgreater back into your M.I.T. slacks and catch up on your Woody Allen. \\
\textgreater \textgreater \\
\textgreater This is an outrage!  I don't even own a dog. \\
\textgreater \\
Of course you do.  You married it a while ago, remember? \\ \hline
\end{tabular}
    \label{tab:twenty_newsgroups2}
\end{table}


\begin{figure}[h!]
    \centering
    \includegraphics[width=0.11\textwidth]{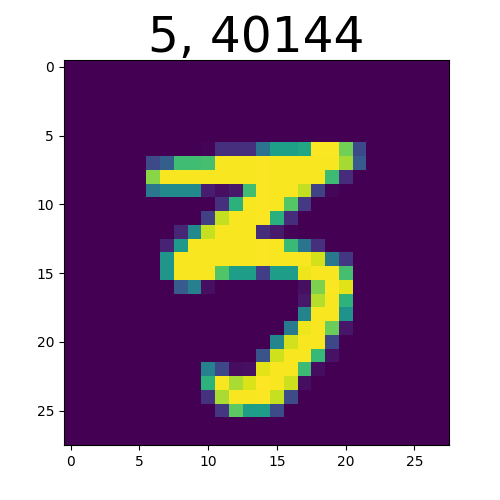}
    \includegraphics[width=0.11\textwidth]{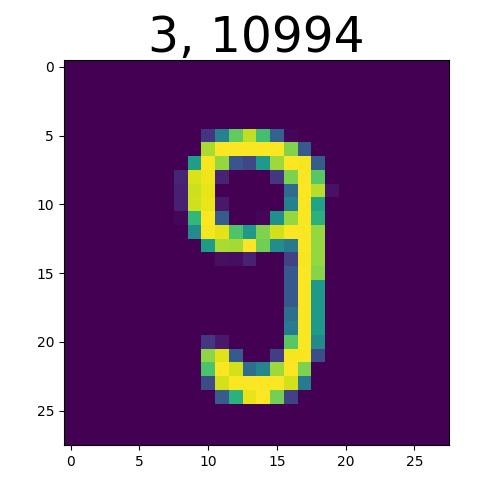}
    \includegraphics[width=0.11\textwidth]{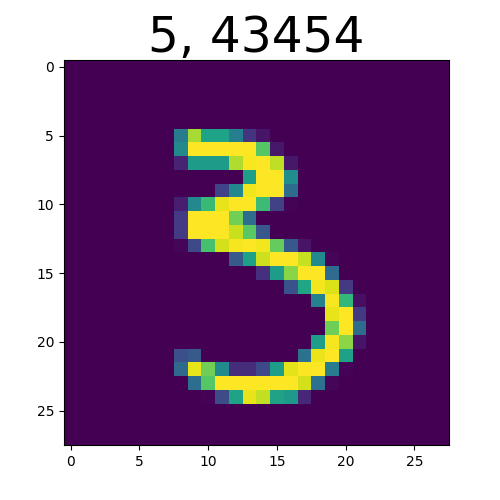}
    \includegraphics[width=0.11\textwidth]{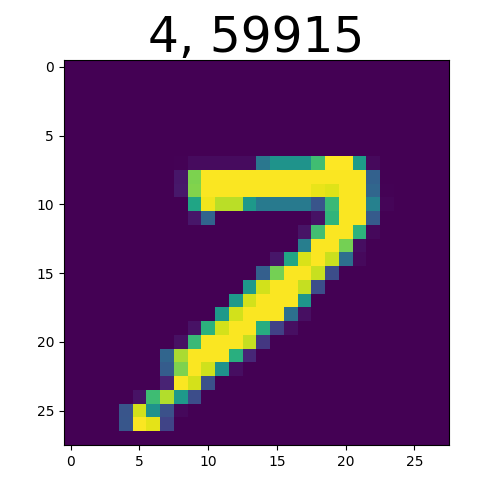}
    \caption{Four mislabeled instances in the MNIST training set. The headings indicate the original label and the index of the image.}
    \label{fig:MislabeledMNIST}
\end{figure}

\begin{table}[htbp]
    \centering
    \caption{Questionable instances in CIFAR-100 for $\alpha = 0.3\%$. This dataset seems to be labeled quite accurately, as we could identify only seven instances in the training set where the image's label and content would not conform with the content.}
    {\footnotesize
    \begin{tabular}{|r|c|}
    \hline
         \textbf{Instance} & \textbf{original label}\\
         \hline
         6093 & crab\\
         24900 & cloud\\
         33823 & television\\
         31377 & camel\\
         48760 & motorcycle\\
         31467 & shark\\
         45694 & forest\\
         \hline
    \end{tabular}}
    \label{tab:CIFARmislabeled}
\end{table}

\begin{table}[htbp]
    \centering
    \caption{A subset of questionable instances in Fashion-MNIST for $\alpha = 0.3\%$. Among the 180 pictures returned by our tool set (we have chosen $\alpha = 0.003$), we have identified 64 instances. 
Hence, more than 35,5\% of the reviewed images were indeed mislabeled. }
    {\footnotesize
    \begin{tabular}{|r|c|c|}
    \hline
         \textbf{Instance} & \textbf{original label} & \textbf{suggested label} \\
         \hline
         3415 & pullover & coat\\
         28264 & dress & shirt\\
         29599 & pullover & top/shirt\\
         37286 & top/shirt & shirt\\
         36049 & shirt & dress\\
         9059 & shirt & dress OR coat\\
         18188 & shirt & dress OR coat\\
         1600 & dress & top/shirt\\
         34381 & shirt & dress\\
         22750 & top/shirt & shirt\\
         39620 & sneaker & ankle boot\\
         50494 & shirt & dress OR coat\\
         38208 & ankle boot & sneaker\\
         53257 & top/shirt & shirt\\
         29487 & shirt & dress\\
         13026 & shirt & dress\\
         20544 & shirt & dress\\
         51464 & top/shirt & shirt\\
         28764 & pullover & top/shirt\\
         29154 & shirt & dress\\
         24804 & top/shirt & shirt\\
         28341 & top/shirt & dress\\
         46125 & pullover & top/shirt\\
         46259 & dress & coat ?\\
         25419 & top/shirt & shirt\\
         36325 & shirt & ?\\
         29728 & coat & ?\\
         43703 & top/shirt & ?\\
         45536 & pullover & top/shirt\\
         3512 & top/shirt & dress\\
         22264 & top/shirt & dress\\
         4027 & shirt & dress\\
         33982 & coat & top/shirt\\
         17243 & shirt & dress\\
         34804 & pullover & top/shirt\\
         20701 & pullover & dress\\
         55829 & dress & top/shirt\\
         35505 & dress & shirt\\
         36061 & dress & shirt\\
         38722 & top/shirt & shirt\\
         33590 & top/shirt & shirt\\
         44903 & top/shirt & dress\\
         50013 & shirt & dress\\
         40513 & pullover & top/shirt\\
         46926 & top/shirt & dress\\
         21771 & shirt & top/shirt\\
         1074 & shirt & top/shirt\\
         42018 & pullover & dress\\
         42110 & dress & pullover\\
         51735 & top/shirt & dress\\
         \hline
    \end{tabular}}
    \label{tab:FashionMNISTmislabeled}
\end{table}

\vspace{12pt}

\end{document}